\pdfoutput=1

\documentclass[11pt]{article}

\usepackage[final]{acl}

\usepackage{times}
\usepackage{latexsym}
\usepackage{amssymb}
\usepackage{amsmath}
\usepackage{booktabs}
\usepackage{multirow}
\usepackage{float}

\usepackage[T1]{fontenc}

\usepackage[utf8]{inputenc}

\usepackage{microtype}

\usepackage{inconsolata}

\usepackage{graphicx}

\title{Contrastive Distillation of Emotion Knowledge from LLMs for Zero-Shot Emotion Recognition}

\author{
  Minxue Niu \quad Emily Mower Provost \\
  University of Michigan, Ann Arbor, Michigan, USA \\
  \texttt{\{sandymn, emilykmp\}@umich.edu}
}

\begin{document}
\maketitle
\begin{abstract}
The ability to handle various emotion labels without dedicated training is crucial for building 
adaptable Emotion Recognition (ER) systems. Conventional ER models rely on training using fixed label sets and struggle to generalize beyond them. On the other hand, Large Language Models (LLMs) have shown strong zero-shot ER performance across diverse label spaces, but their scale limits their use on edge devices. 
In this work, we propose a contrastive distillation framework that transfers rich emotional knowledge from LLMs into a compact model without the use of human annotations. We use GPT-4 to generate descriptive emotion annotations, offering rich supervision beyond fixed label sets. By aligning text samples with emotion descriptors in a shared embedding space, our method enables zero-shot prediction on different emotion classes, granularity, and label schema. The distilled model is effective across multiple datasets and label spaces, outperforming strong baselines of similar size and approaching GPT-4's zero-shot performance, while being over 10,000 times smaller. 
\end{abstract}

\section{Introduction}

\begin{figure}[h]
  \centering
  \includegraphics[width=0.8\linewidth]{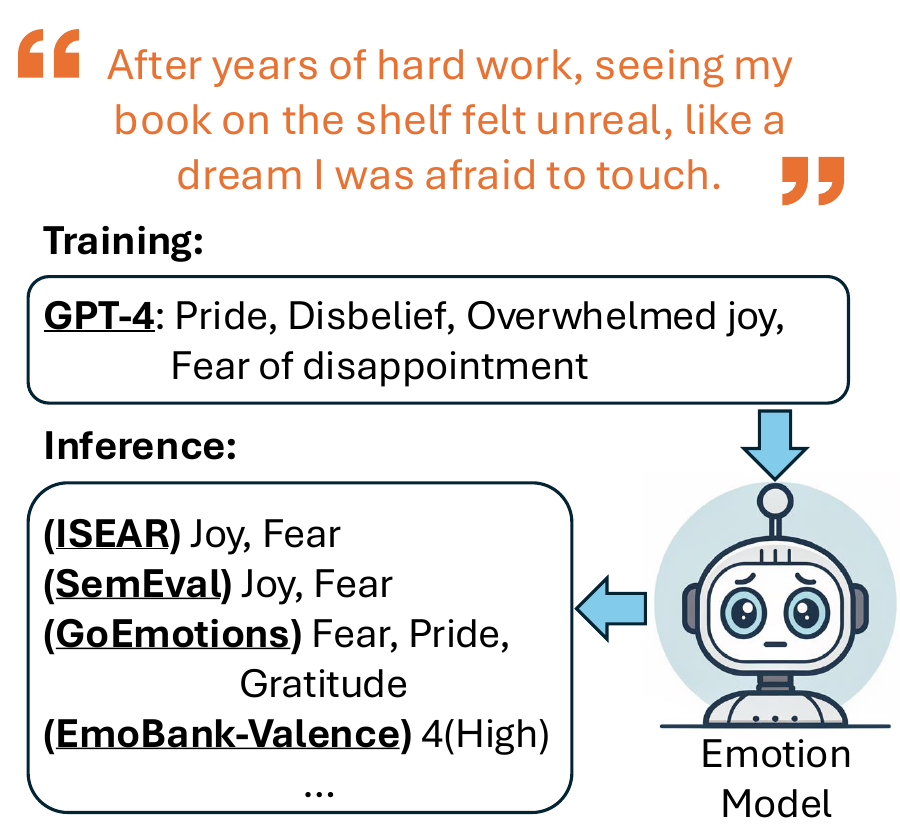}
  \caption{Our model is trained with rich emotion descriptors generated by GPT-4. During inference, this much smaller model can flexibly perform classification or regression on new label spaces.}
  \label{fig:overview}
\vspace{-10pt}
\end{figure}
Emotion Recognition systems are increasingly integral to applications such as mental health monitoring~\cite{trotzek2018utilizing, liu2024affective}, customer service~\cite{plaza2022emotion}, and human-computer interaction~\cite{cowie2001emotion}. There is a growing need for models that can flexibly adapt to application-specific requirements. Zero-shot ER, the ability of models to generalize to new text domains and label spaces without extra annotation or retraining, has attracted growing research interest~\cite{zhan2019zero, Olah2021-or, lian2024gpt}. While significant progress has been made on generalizing across text domains~\cite{chochlakis2023using,feng2020review}, zero-shot inference on new emotion label spaces (which may include unseen labels) remains an arguably more underexplored and challenging problem~\cite{chochlakis2023using}. In this work, we propose a compact, generalizable framework that learns emotion representations from LLM supervision and supports flexible zero-shot prediction across diverse emotion taxonomies.

Different applications often involve different emotional expressions and recognition needs. For instance, customer service systems often focus on identifying coarse-grained emotions such as positive, neutral, or negative emotions to understand customer experiences~\citep{lee2005toward, yurtay2024emotion}, while mental health monitoring can benefit from more fine-grained emotional signals to capture subtle psychological states~\citep{trotzek2018utilizing, rasool2025nbert}.
Consequently, ER datasets have been developed to meet different needs. Common design choices include \textbf{label schema} (e.g., whether to use categorical or dimensional labels), \textbf{label granularity} (whether to adopt an extensive set of fine-grained emotions or a smaller set of distinct categories), and \textbf{label selection} (which specific emotions or dimensions to include in the label set)~\citep{demszky2020goemotions, rashkin2019towards, wallbott1986universal, Buechel2017-zn}.
While models can be trained to 
support multiple label spaces in multitask setups~\citep{vu2021multitask}, they are not designed to handle novel labels at inference. As a result, dedicated data collection and retraining is needed for each new application, limiting these models' use across various downstream tasks.

Recent advances in LLMs have opened new possibilities for ER. Through text interactions, LLMs can reason about emotions and their causes~\cite{wang2023emotional, zhang2024sentiment}. They achieve strong zero-shot performance on various emotion spaces, especially on large, fine-grained ones~\citep{liu2024emollms, niu2024text}. Moreover, they can generate rich descriptions of emotion states, offering a level of expressiveness beyond the constraints of traditional fixed label spaces~\cite{niu2024text, Bhaumik2024-xn}. 

However, several factors limit the direct use of LLMs in applications. First, 
leading LLMs can contain trillions of parameters, making it infeasible to deploy them on personal 
devices~\cite{zheng2025review}. Yet, many good or even state-of-the-art ER models rely on much smaller architectures ~\citep{al2024challenges, ameer2023multi}, suggesting that extensive computational overhead may not always be necessary.
Additionally, the powerful language representation capability of LLMs enables them to retain rich linguistic content, which can raise privacy concerns in sensitive settings~\cite{yao2024survey}. 
Finally, LLMs typically generate free-form text rather than structured emotion representations, requiring additional post-processing to extract usable emotion labels~\cite{xia2024fofo}.

This work proposes a method to build \textbf{compact models that enable zero-shot ER across various label spaces without additional human supervision and without additional training.}
As illustrated in Figure \ref{fig:overview}, we propose a novel approach that distills emotion knowledge from GPT-4 into a small, BERT-based model. Instead of relying on fixed categorical labels, we prompt GPT-4, as a representative leading LLM, to generate nuanced, free-form emotion descriptors that offer rich supervision beyond predefined label sets. 
We train our model using a contrastive learning framework inspired by CLIP~\citep{radford2021learning}, to align text samples and emotion terms in a shared latent emotion embedding space. Our model shows good zero-shot performance across multiple datasets with diverse label spaces. It outperforms strong baseline models of comparable size and approaches GPT-4 performance on multi-label emotion classification while being much smaller. Finally, we perform a nearest-neighbor analysis to show that our model learns meaningful clustering of semantically related emotions, enhancing the interpretability of our results.

In summary, we make the following contributions: 1) We introduce a method for generating rich, descriptive emotion annotations using GPT-4. 
2) We propose a contrastive framework that extends CLIP to multi-label tasks, enabling zero-shot ER across variable label spaces without additional human annotation. 3) We evaluate our model against multiple supervised and zero-shot baselines, showing strong performance with a compact model size. Our data and code are released at \url{https://github.com/chailab-umich/uni_emotion_release}.

\section{Related Work}

\subsection{Zero-shot ER Models}
\label{sec:related-zs}

Zero-shot ER has attracted broad research interest. 
Most existing work focuses on generalizing to new text domains, where common strategies include using encoders trained on more diverse texts~\citep{chochlakis2023using}, domain adaptation and adversarial learning~\citep{feng2020review, gao2023adversarial}, and data augmentation~\citep{bo2025toward}. However, these approaches assume a fixed label space between training and testing.

An arguably more challenging setup is generalization to unseen label spaces. Earlier approaches have used soft labels on a set of seen emotions as features to predict unseen ones, effectively learning correlations between emotion labels~\citep{tesfagergish2022zero, buechel2021towards}. These approaches are typically applied to coarse-grained emotion classes and often benefit from few-shot tuning. Similarity-based methods are also popular. For example, \citet{Olah2021-or} proposed to obtain semantic embeddings of the emotion label and text input and use cosine similarity to measure their alignment. \citet{stanley2023emotion} applies a similar approach to speech ER. Another line of work is entailment-based ER, which frames the task as predicting sentence entailment using crafted input like ``this text expresses [emotion]''~\citep{yin2019benchmarking, del2022natural, bareiss2024english}. In general, zero-shot ER remains highly challenging: many studies report a gap of 0.3-0.5 in F1 between finetuned and zero-shot setups~\citep{Olah2021-or, del2022natural, chochlakis2023using}. 

\subsection{ER with LLMs}
\label{sec:related-llm}
With their growing 
Natural Language Understanding (NLU) capabilities, LLMs have been evaluated as an alternative for zero-shot ER. Several studies have shown that LLMs are capable of interpreting emotions and reasoning about their causes~\citep{wang2023emotional, zhao2023chatgpt, Tak2023-fk}. LLMs also demonstrate strong zero-shot performance across various label spaces, approaching finetuned smaller models or even human performance~\citep{liu2024emollms, tak2024gpt, niu2024text}. Despite their promise, LLMs produce free-text outputs rather than structured emotion predictions. As a result, existing methods rely on prompt engineering to elicit responses in a specific format, requiring extra processing to extract usable emotion representations. Moreover, due to their extreme scale, they are not practical for deployment on edge devices or in resource-constrained environments~\citep{zheng2025review}.

\subsection{Descriptive Emotion Labels}
\label{sec:related-descriptive}
Categorical emotion label spaces face critiques for 
not being flexible enough to capture the full nuance of human emotions~\cite{cowie2003describing}. In addition to expanding label granularity in predefined taxonomies~\citep{demszky2020goemotions, rashkin2019towards}, recent studies have proposed using 
unrestricted generative labels as an alternative, both in human annotation and in ER model outputs~\citep{hoemann2024using,stanley2023emotion, bhaumik2024towards, niu2024text}. Generative descriptions enhance label richness and accuracy, but there is no standardized way to map open-ended textual descriptions to existing label sets or apply them in downstream applications.


\section{Methods}
\subsection{Obtaining Rich Emotion Annotations}
\label{sec:method-prompt}
We adopt a zero-shot prompting strategy following previous work~\cite{niu2024text} to encourage GPT-4 to generate multiple descriptive terms (words or phrases) that capture the nuanced emotion states conveyed in the text. We use the following system prompt to obtain annotations for the GoEmotions dataset (see Section~\ref{sec:data}): \textit{You are an emotionally-intelligent and empathetic agent. You will be given a piece of text, and your task is to identify the emotions expressed by the writer of the text. Reply with only the emotion descriptors (words or phrases), separated by commas. If no emotion is clearly expressed, reply with ``neutral''.}

We use the ``gpt-4-1106-preview'' version of GPT-4 deployment through the Azure API. Across 43.4k samples in the GoEmotions training split, GPT-4 generated $2,173$ unique emotion terms, in contrast to $27$ labels in human annotations. Each sample has $1.91$ (standard deviation $0.65$) terms on average and $13.0$ (standard deviation $5.9$) characters per term. We show the human classification and GPT-4 generated labels on 10 randomly selected samples in Appendix A. Upon manual inspection, we notice that most terms in human labels also occur in GPT-4 generations, while GPT-4 tends to produce more complex or less common terms, such as ``reminiscence'' and ``foreboding''. It also uses extent descriptors like ``faint optimism'' and ``a little bummed'', or specifies emotional nuances or causes, e.g., ``fear of abandonment'', ``fear of embarrassment'', and ``fear of failure''. Finally, the format of the generated terms is not normalized; e.g., it includes a mixture of nouns (``fear'') and adjectives (``fearful'') for the same emotion.  We intentionally retain this variability, hypothesizing that exposure to diverse expressions may improve the generalizability of the model.

\begin{figure*}[htbp]
\vspace{-10pt}
  \centering
  \includegraphics[width=1.03\linewidth]{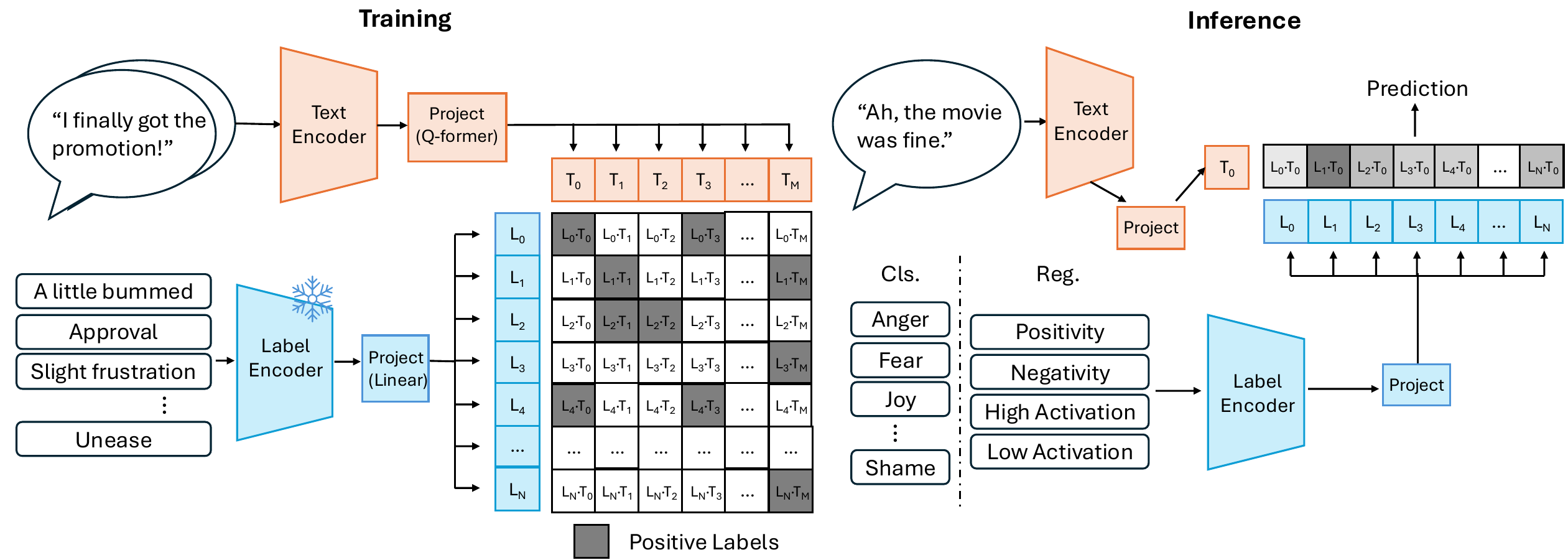}
  \caption{Overview of our contrastive distillation model structure.}
  \label{fig:model}
\end{figure*}

\subsection{Contrastive Distillation}
\label{sec:method-contrastive}

Our model structure is 
illustrated in Figure~\ref{fig:model}. During training, a batch of text samples is encoded into hidden representations using a text encoder. We collect all the emotion labels associated with the texts in the batch and embed them through a separate label encoder. We adopt pretrained BERT models for both the text and label encoders, keeping the label encoder frozen during training. BERT is chosen due to its strong text representation capabilities among models of similar sizes (around 100 million parameters). Also, BERT-based models are frequently used in zero-shot ER baselines (\citet{Olah2021-or, del2022natural, bareiss2024english}
, making them a strong and representative base model for our experiments.

Then, two separate projection modules transform the text and label embeddings into the same space, which we consider as our emotion representation space. On the text side, we adopt a Q-Former~\citep{li2023blip} to capture salient emotional information from the contextualized word representations. On the label side, since labels are relatively short, we apply a simple linear projection to their [CLS] token embedding~\citep{devlin2019bert}.  We denote $\mathbf{T} \in \mathbb{R}^{B \times d}$ as the text embeddings after normalization, and $\mathbf{L} \in \mathbb{R}^{N \times d}$ as the label embeddings. B is batch size, N is the number of labels in that batch (N varies across batches). The dimension of the learned emotion space $d$ can be tuned as a hyperparameter. The model outputs an alignment matrix $\mathbf{M} = \mathbf{T} \mathbf{L}^{\top} / \tau$. Since $\mathbf{T}$ and $\mathbf{L}$ are both normalized, the dot product returns the cosine similarity for each text-label pair. $\tau$ is a temperature scalar. We set $d = 768$ for fair comparison with baseline models (Section \ref{sec:baselines}), and we set $\tau = 0.1$ .

We modify the loss function in CLIP to enable multi-label training.
We first apply a sigmoid function to the alignment matrix to obtain the predicted probabilities. Then, using the ground-truth binary label alignment matrix $\mathbf{Y} \in \{0, 1\}^{B \times N}$, we aim to maximize the log probability for matched (positive) text-label pairs, while minimizing the log probability for unmatched (negative) pairs. The resulting contrastive sigmoid loss is defined as: 
\vspace{-20pt}

\begin{equation*}
\begin{split}
\mathcal{L} = &- \frac{1}{\sum \mathbf{Y}} \sum_{i=1}^{B} \sum_{j=1}^{N} \mathbf{Y}_{ij} \log(\sigma(\mathbf{M}_{ij})) - \\& \frac{1}{\sum (1 - \mathbf{Y})} \sum_{i=1}^{B} \sum_{j=1}^{N} (1-\mathbf{Y}_{ij}) \log(1-\sigma(\mathbf{M}_{ij}))
\end{split}
\end{equation*}
where $\sigma(\cdot)$ denotes the sigmoid function. The alignment matrices are generally very sparse, so we average the log probability for positive and negative pairs separately before adding them together to balance the precision and recall. 

\subsection{Zero-shot Inference}

We consider three common formulations in text-based ER: single-label classification, multi-label classification, and regression. 

As discussed in Section 3.2, the model is trained to predict the GPT-4 generated emotion labels. 
During inference, our trained model can flexibly adapter to different emotion label spaces by replacing the GPT-4 labels with the fixed set of human labels that are specific to each test dataset, as shown in Figure~\ref{fig:model} (Inference). For single-label classification, we output the label from the fixed set that has the highest probability. For multi-label classification, we apply label-specific thresholds, calibrated on a small validation set (see Section \ref{sec:setup}) and output all labels that are above this threshold. For regression tasks we convert the original valence and activation labels to four discrete categories: ``Positivity'', ``Negativity'', ``High Activation'', and ``Low Activation''. The predicted valence score is then computed as the difference $M[\text{text}, \text{Positivity}] - M[\text{text}, \text{Negativity}]$, and the predicted activation score is computed as $M[\text{text}, \text{High Activation}] - M[\text{text}, \text{Low Activation}]$.

With contrastive distillation, our model learns a compact and reusable emotion space from GPT-4's descriptive supervision, and it can generalize seamlessly to various label spaces.

\section{Experiments}
\subsection{Data}
\label{sec:data}
We use multiple text-based ER datasets for training and evaluation, covering diverse emotion label spaces. The training set is constructed using text samples from the GoEmotions dataset paired with rich descriptive annotations generated by GPT-4 (see Section \ref{sec:method-prompt}). GoEmotions is selected for training due to its diverse emotional content and its fine-grained human annotation schema, which allows a controlled comparison between human and GPT-4 supervision. We evaluate the trained model's zero-shot performance on three additional datasets using their test split, each with distinct label spaces. 

\textbf{GoEmotions. } GoEmotions~\citep{demszky2020goemotions} contains approximately 58k English Reddit comments annotated with one or more of 27 fine-grained emotion classes. 
For training, we use only the raw text from the training split ($N=43.4$k), paired with GPT-4 generated descriptive emotion labels (not the original human annotations).

\textbf{SemEval. } We use the data from SemEval-2018 Task 1: Affect in Tweets~\citep{Mohammad2018-mh}. It also provides \textit{multi-label} categorical emotion annotations. This setup is most similar to GoEmotions, but it only has 11 emotion classes. 

\textbf{ISEAR. }The International Survey on Emotion Antecedents and Reactions dataset~\citep{wallbott1986universal} contains short self-reported descriptions of emotional experiences collected across different countries. Each instance is labeled in a \textit{single-label} manner with one of seven emotions.

\textbf{Emobank. } EmoBank~\citep{Buechel2017-zn} is the only dataset we use with dimensional emotion annotations. It consists of approximately 10k English sentences from a variety of genres, including news articles, blogs, and fiction. Each sample is annotated by readers along three dimensions on a 5-point scale. In this study, we focus on predicting valence and activation scores. 
\begin{table*}[t]
\centering
\small
\begin{tabular}{l|c|c|c|cc|cc}
\toprule
\multirow{2}{*}{Model} & GoEmotions & SemEval & ISEAR & \multicolumn{2}{c|}{EmoBank-V} & \multicolumn{2}{c}{EmoBank-A} \\
& macro-F1  & macro-F1  & macro-F1  & Rho ($\rho$) & PCC & Rho ($\rho$) & PCC \\
\midrule
BERT-FT         & \underline{0.477}  & \underline{0.563}  & 0.677  & \underline{0.764} & \underline{0.790} & \underline{0.470} & \underline{0.556} \\
GPT-4-ZS        & 0.319  & 0.486  & \underline{0.728}  & 0.678 & 0.718 & 0.369 & 0.324 \\
\midrule
BERT-ZS & /  & 0.205 & 0.336  & -0.033 & -0.004 & -0.011 & -0.022 \\
Similarity     & 0.137 & 0.391  & \textbf{0.504}  & 0.428 & 0.401 & 0.110 & 0.119 \\
Entailment     & 0.073  & 0.430  & 0.222  & 0.534 & 0.485 & 0.102 & 0.105 \\
\textbf{Ours}           & \textbf{0.299$\pm$0.005}  & \textbf{0.480$\pm$0.003}  & 0.479$\pm$0.015  & \textbf{0.613$\pm$0.008 } & \textbf{0.593$\pm$0.003} & \textbf{0.155$\pm$0.028} & \textbf{0.202$\pm$0.040} \\
\bottomrule
\end{tabular}
\vspace{-3pt}
\caption{Zero-shot performance across four datasets. For our model, we train the model with five random seeds and we report the mean and standard deviation of the performance metrics. The best overall performance is underlined. The best zero-shot performance among comparably sized models is highlighted in bold.}
\label{tab:results-main}
\end{table*}

\subsection{Baseline Models}
\label{sec:baselines}


\paragraph{Upper-bound models. }
 We contextualize our model's performance with two upper-bound baselines. First, we evaluate \textbf{GPT-4 zero-shot (GPT-ZS)} performance by prompting GPT-4 to perform zero-shot ER. The prompts are adapted from prior work~\cite{niu2024text}, formatted similarly to our generation prompts (Section~\ref{sec:method-prompt}) but modified to elicit dataset-specific outputs. Given that GPT-4 is over 10k times larger than BERT\footnote{Estimate based on public parameter counts: GPT-4 >100B vs. BERT-base ~110M}, and our model is distilled from its outputs, we consider GPT-ZS as a performance upper bound. Second, we include a \textbf{finetuned BERT model (BERT-FT)}, which shares the same backbone as our model but is trained with direct supervision. We train a separate model for each evaluation dataset and select the best checkpoint based on validation performance (macro-F1 for classification; PCC for regression).

\paragraph{Baseline Models} 
We compare our model against three baseline models. We use BERT as the base model across the baselines and our methods. The first is a BERT model finetuned on GoEmotions and directly applied to other datasets without further adaptation (\textbf{BERT-ZS}). For classification tasks (SemEval and ISEAR), we reuse the prediction heads for overlapping emotion labels and randomly initialize new heads for unseen labels. For regression tasks (EmoBank), the prediction heads are entirely randomly initialized. This serves as a conservative and relatively weak baseline.
We also include two strong zero-shot baselines with comparable model sizes from prior work
. The first is a \textbf{similarity-based} method~\cite{Olah2021-or}, where cosine similarity is computed between the embedding of the input text and each emotion label augmented with its definition. The second is an \textbf{entailment-based} method~\cite{bareiss2024english}, where a BERT model is finetuned on multiple NLU tasks and predicts the plausibility of the crafted input ``[text] This text expresses [emotion].''

\subsection{Experimental Setup and Metrics}
\label{sec:setup}
\textbf{Training. }
To ensure that no validation or test samples from GoEmotions are seen during the training of our model, we reserve 20\% of GoEmotions training set for model selection. Additionally, to assess model fit on the training task itself, we report micro-averaged F1 scores on the GoEmotions test set. (Macro-averaging is not feasible due to the extensive number of training labels.)

\textbf{Evaluation. }
For classification datasets, we evaluate performance using the macro-averaged F1-score to reflect overall classification effectiveness across classes.
For regression tasks, we report Pearson correlation coefficient (PCC) to measure the linear relationships between predicted and ground-truth scores, and Spearman’s rank correlation coefficient ($\rho$) to assess monotonic relationships.
For EmoBank, we report regression performance separately for valence and activation dimensions.

It is important to note that comparisons on the GoEmotions test set are not fully controlled, as models differ in the additional information available during training: our model is exposed to the text content in the GoEmotions training set, while the entailment-based baseline is finetuned on a broad range of NLU tasks, including some emotion-related ones. The similarity-based baseline relies on emotion label definitions from WordNet. We still report their performance as zero-shot since none of the models has seen the GoEmotions human annotations or test set text samples, but we note that the results should be interpreted with care.

For multi-label classification tasks, we perform an additional threshold calibration step on a validation set for our model and the baseline methods. Specifically, we determine the optimal threshold for each emotion class by searching between 0 and 1 (in increments of 0.05), selecting the value that yields the best performance on the validation split of each dataset. While this setup is not strictly zero-shot, it is consistent with prior zero-shot SER work~\cite{Olah2021-or}. We believe this step is necessary because of the nature of multi-label classification: in this setup, the goal is not to output the most likely emotion, but instead to identify all emotions that are present.  In this case, a threshold is needed to make this judgment. We further discuss this limitation in the Limitation Section. The regression and single-label classification tasks are evaluated in a strictly zero-shot manner.

For all baseline models, we report results based on our own replication to ensure consistency across datasets and evaluation setups. Our replicated results are comparable to those reported in the original papers. Additional training and hyperparameter details can be found in our released code.

\vspace{-3pt}
\section{Results}
\vspace{-3pt}
\subsection{Overall Performance}

We first compare the zero-shot performance of our model against all baseline methods in Table~\ref{tab:results-main}. 

First, comparing the two upper-bound models, we find that dataset-specific fine-tuning provides significant benefits: BERT-FT achieves the best performance on most benchmarks.
Although GPT-4 is substantially larger, it only outperforms BERT-FT on the ISEAR dataset. This is possibly due to the specificity of emotion datasets: label sets, annotation instructions, and cultural or contextual assumptions can vary widely, making it difficult even for powerful general-purpose models to fit specific emotion distributions without adaptation.

Our model shows competitive performance on multi-label classification tasks. It outperforms both strong zero-shot baselines and approaches GPT-4 performance (e.g., $0.486$ for GPT-4 vs. $0.480$ for ours on SemEval). Although our model has been exposed to GoEmotions training texts (but not the label space, see Section \ref{sec:setup}) and may be more familiar with the text domain, the substantial performance margin over baselines suggests a genuine ability to generalize to new label spaces. This is further validated on SemEval, where all models are fully zero-shot, and ours achieves $0.480$ F1, outperforming $0.391$ for the similarity-based method and $0.430$ for the entailment-based method.
We observe that the performance of entailment-based method varies substantially across datasets (e.g., strong on SemEval but nearly-random on GoEmotions), likely due to their sensitivity to domain shifts and prompt formulations~\citep{yin2019benchmarking}.

Our model performs slightly worse than the similarity-based baseline on the single-label ISEAR dataset (macro-F1: Ours $0.479$, Similarity $0.504$, Entailment $0.222$). We suspect this is due to our model’s multi-label training setup, which encourages capturing multiple plausible emotions rather than selecting the most dominant one. For instance, it often confuses guilt and shame, which do naturally co-occur. 
Further analysis and targeted experiments may help clarify this behavior or improve the model for single-label scenarios.

Finally, although not explicitly trained for regression tasks, ours and both baseline models achieve surprisingly strong results on valence regression. Our model obtains a Spearman correlation of $0.613$ and a PCC of $0.593$, outperforming $0.534$ $\rho$ and $0.48$5 PCC for the entailment-based model, and $0.428$ $\rho$ and $0.401$ PCC for the similarity-based model.
Performance on activation prediction is notably lower across all models. Activation is generally more difficult to infer from text alone~\citep{Buechel2017-zn, wagner2023dawn}, and activation-related terms such as ``high activation'' or ``low activation'' are less commonly included in both human-annotated labels and GPT-4-generated descriptions.
Overall, our approach shows encouraging results for zero-shot regression tasks, but further research is needed to close its performance gap with supervised models.

\subsection{Ablation Studies}
Since the standard deviation of performance metrics was found to be small during training (see Table~\ref{tab:results-main}), we use a fixed random seed (42) for all ablation studies to reduce computational cost.

\begin{table}[t]
\centering
\small
\begin{tabular}{c|c|ccc}
\toprule
\textbf{Dim} & \textbf{Train} & \multicolumn{3}{c}{\textbf{GPT-4 distillation}} \\
\cline{3-5}
& & G & S & I \\
\midrule
50  & 0.428 & 0.274 & 0.477 & 0.451 \\
100 & 0.443 & 0.290 & 0.475 & 0.449 \\
200 & 0.447 & 0.293 & 0.479 & 0.486 \\
768 & \textbf{0.454} & \textbf{0.296} & \textbf{0.476} & \textbf{0.488} \\
\bottomrule
\end{tabular}
\caption{Comparison of emotion space dimensionality. ``Train'' column shows the performance on GoEmotions test set with GPT-4 labels, measured by micro-F1. ``G'', ``S'', and ``I'' refer to performance on GoEmotions, SemEval, and ISEAR respectively, measured by macro-F1.}
\label{tab:dim-comparison}
\end{table}

\textbf{Dimension Size.} We investigate the impact of the emotion space dimension $d$. In the 
Table~\ref{tab:results-main}, we use $d=768$ to maintain consistency with the baseline models for fair comparison. However, 
smaller $d$ is desirable for downstream applications due to lower computational costs and smaller model sizes.

To explore this trade-off, we conducted additional experiments with $d \in \{200, 100, 50\}$ across the classification datasets. As shown in Table~\ref{tab:dim-comparison}, performance generally drops slightly as the dimension decreases, but the decline remains small even under aggressive reductions (e.g., $d=50$). This suggests that our model can maintain strong performance even with a compact emotion space, making it suitable for resource-efficient applications.

\begin{table}[t]
\centering
\small
\begin{tabular}{l|c|ccc}
\toprule
\textbf{} & Train & G & \textbf{S} & \textbf{I} \\
\midrule
Ours (GPT-4 labels) & 0.454 & 0.296 & 0.476 & 0.488 \\
\midrule
\multicolumn{5}{c}{\textbf{Human labels}} \\
\midrule
overall & 0.587 & 0.475 & 0.414 & 0.410 \\
seen classes & / & 0.475 & 0.509 & 0.488 \\
unseen classes & / & / & 0.161 & 0.215 \\
\bottomrule
\end{tabular}
\caption{Comparison of models trained with GPT-4 generated labels versus human-labels. For models trained with human labels, we also report separate results on classes seen in the training dataset (8 out of 11 in SemEval, 5 out of 7 in ISEAR). Note that GoEmotions performance under human supervision reflects a supervised setting, while all others are zero-shot evaluations.}
\label{tab:human-compare}
\end{table}


\textbf{GPT-4 Supervision.} We next probe the effect of using GPT-4 generated labels for supervision. We compare our model with a variant trained on the same samples but supervised with human annotations, rather than GPT-4 generated annotations. As Table~\ref{tab:human-compare} shows, models trained on human labels perform better on the GoEmotions dataset itself\textemdash as expected, due to direct supervision\textemdash but exhibit worse zero-shot generalization to SemEval and ISEAR. The model trained on human labels performs well on seen classes, even outperforming our GPT-4-distilled model on SemEval, but its performance drops sharply on unseen classes, with macro-F1 scores of only $0.161$ for SemEval and $0.215$ for ISEAR. Yet, both models share the same contrastive architecture, making it theoretically possible for the human-supervised model to generalize to unseen labels because of BERT's existing semantic embedding space. However, the richness of the supervision makes a substantial difference. Notably, GoEmotions already provides one of the most extensive categorical label sets among text-based ER datasets. These results suggest that it remains difficult to learn a generalizable emotion space from a fixed and limited set of labels, underscoring the advantage of distilling from rich, descriptive annotations.

\subsection{Emotion Space Probing}

\begin{table*}[t]
\centering
\small
\begin{tabular}{c|c|c|c}
\toprule
\textbf{Target Emotion} & \textbf{BERT (768D)} & \textbf{Ours (768D)} & \textbf{Ours (50D)} \\
\midrule
Admiration & Sympathy, Gratitude & Reverence, Amazement & Accomplishment, Reverence \\
Gratitude & Satisfaction, Admiration & Appreciation, Grateful & Appreciation, Grateful \\
Approval & Disapproval, Recommendation & Positive Surprise, Positive & Positive Interest, Positive Surprise \\
Annoyance & Distraction, Reluctance & Irritation, Annoyed & Irritation, Annoyed \\
Curiosity & Suspicion, Surprise & Interest, Intrigue & Interest, Slight Hopefulness \\
\bottomrule
\end{tabular}
\caption{Top-2 most similar GPT-4-generated emotion terms retrieved for each of the five most frequent emotion labels in the GoEmotions test set, using BERT text embeddings, our full model, and a reduced 50D version.}
\label{tab:emotion-retrieval}
\end{table*}

Finally, to interpret the learned emotion space, we examine its nearest-neighbor structure. We use all 27 classes from the GoEmotions dataset as target emotions and GPT-4-generated emotion description terms on its test split as the candidate pool ($N = 684$). We only use the test set to ensure that the retrieved structure reflects generalization rather than overfitting to the training data.

We compare our model against a BERT encoder baseline. For BERT, we extract the [CLS] token embeddings of all emotion labels/terms. For our model, we encode these terms using our trained Label Encoder and Projector (as shown in Figure \ref{fig:model}). Since the Label Encoder is initialized with the BERT encoder and frozen during training, our encoder only differs from BERT by one linear layer. For each target emotion, we retrieve the top-4 most similar terms from the candidate pool using cosine similarity. Due to space constraints, Table~\ref{tab:emotion-retrieval} shows the five most frequent target emotions and their top-2 neighbors; full results are provided in Appendix B.

Manual inspection suggests our model retrieves more emotionally aligned neighbors, compared to BERT. For instance, for ``Admiration'', our model returns ``Reverence'' and ``Amazement'', whereas BERT returns ``Sympathy'' and ``Gratitude''. We also observe that BERT tends to prioritize part-of-speech consistency, e.g.,  failing to retrieve ``annoyed'' for ``annoyance'' or ``grateful'' for ``gratitude''.  In some cases, it even retrieves antonyms such as ``Disapproval'' for ``Approval''. These behaviors are likely due to semantic relatedness in the general language space, but are unfavorable for emotion-specific use cases. Our contrastive training helps mitigate these effects. Additionally, we conduct the same experiment using our 50D model, aggressively compressing the emotion space. The retrieved neighbors remain largely consistent with those from the 768D model and, in our judgment, still align more closely with the intended target emotions than BERT. These results demonstrate that our approach can learn a more efficient representation space that better preserves emotion nuances compared to language-focused embeddings. 

\section{Discussions}
\label{sec:discussion}
Zero-shot generalization is a highly desirable capability for ER systems, as it enables flexible adaptation to applications without the need for extra data collection or retraining. In this work, we design a compact model that distills emotional knowledge from GPT-4, achieving zero-shot generalization without the prohibitive scale of LLMs. 

Our results are encouraging. First, we demonstrate that our contrastive learning framework enables the model to handle diverse ER setups, including both multi- and single-label classification, as well as regression. 
Second, through a nearest-neighbor retrieval analysis, we show that our model captures emotional saliency from general language representations, and this structure remains largely preserved even when the embedding dimensionality is reduced from $768$ to $50$. Together, these results suggest that our approach yields compact, generalizable emotion representations that can be readily applied to a variety of downstream tasks. In settings where further (few-shot) tuning is possible, these representations provide a strong and efficient starting point for adaptation.

Our results also invite reflection on what makes an effective representation space for emotions. While LLMs demonstrate strong emotion understanding (Section \ref{sec:related-llm}), they operate in the full language space, which encodes a wide range of information beyond emotion. As a result, they can be larger and more resource-intensive than necessary for emotion tasks. In contrast, our approach distills a dedicated emotion space that focuses on emotionally salient features, showing that strong emotion understanding can be achieved with significantly less computation. We hope these findings encourage future research in this domain.


%

\section{Conclusion}
We present a contrastive distillation framework that extracts emotional knowledge from LLMs into a compact, BERT-sized model. Our method learns to map both text inputs and GPT-4 generated label descriptors into a joint representation space without the need of human annotations, and it enables zero-shot prediction across diverse label sets and task types. Experiments show strong performance across datasets and label spaces, outperforming comparable zero-shot baselines and approaching GPT-4 zero-shot performance while remaining far smaller in size. We discuss practical directions to further improve the model, as well as potential ethical considerations surrounding emotion annotation and model fairness.

\section*{Limitations}
For zero-shot inference on multi-label classification tasks, we calibrate prediction thresholds on a validation set for both our model and baseline models. We also ran experiments without calibration, where all models showed significant drops in F1 scores, but are still above random baseline and the relative ranking among the models remains the same. We think that threshold calibration is an essential step for multi-label classification under the current setups, where the model independently predicts each label without considering the full label set. However, human judgments are influenced by the full set of available alternatives, For example, in ISEAR, all positive samples are typically labeled as ``joy'', as it is the only available positive category. If given more fine-grained options, annotators may choose more accurate descriptions like ``proud' or ``excited'' while dropping ``joy''. The calibration step serves as weak supervision to help models adjust to these differences. To remove this constraint, future work could explore methods that considers all label options jointly for each sample.

While our framework shows strong potential, there are several practical directions for further improving its generalizability and robustness. First, this work focuses on generalization to unseen labels, while new text domains can also pose challenges. Instead of using GoEmotions as the sole source of text for supervision, future work could incorporate more diverse textual sources to improve generalization across varied contexts. Second, although our model is compatible with multiple emotion label spaces, the current loss design is best aligned with multi-label classification\textemdash where we also observe the strongest empirical performance. Future work could explore multi-stage training strategies to better prepare the model for specific downstream applications while preserving its zero-shot generalization ability.

\section*{Ethical Considerations}
Emotion recognition inherently involves subjectivity, as emotional expressions and interpretations can vary significantly across individuals and cultural backgrounds~\cite{zhang2022examining, scherer2011eye}. As such, bias and fairness are persistent concerns in emotion annotation and modeling~\cite{mao2023biases, zhang2022examining, xu2020investigating}. In our work, we distill emotion supervision from GPT-4. This approach has potential benefits: LLMs like GPT-4 are trained on large-scale, diverse data and are explicitly designed with fairness considerations in mind~\cite{mu2024rule}, which may reduce some forms of annotator-specific bias. However, any representational or linguistic biases in GPT-4 will propagate into the distilled model. In our annotations, we also observe that GPT-4 tends to generate emotion terms that are more complex or infrequent, sometimes diverging from how many people naturally express emotions in everyday settings. While such richness can enhance expressiveness, it may also reflect linguistic or cultural preferences that do not generalize across populations. These patterns underscore the need for careful reflection on the sources and implications of emotion supervision, especially when deploying models in sensitive applications.

\section*{Acknowledgments}

\bibliography{custom}

\clearpage
\onecolumn
\appendix

\section{Examples of human vs. GPT-4 generated labels on GoEmotions}
\begin{table}[H]
\centering
\small
\begin{tabular}{p{7.2cm}|p{2.5cm}|p{4cm}}
\toprule
\textbf{Text} & \textbf{Human Label} & \textbf{GPT-4 Label} \\
\midrule
At least they can make some good pizza & Neutral & Contentment, Satisfaction \\
Embrace the feels my friend. Glad you found happiness even if it is fleeting. & Joy & Supportive, Glad, Accepting \\
Congrats! Vegan baking is still daunting to me but I will conquer it one day! & Gratitude & Encouragement, Determination \\
I AM CALLING THE POLICE & Neutral & Urgency, Fear, Anger \\
Whew lad they're bleeding employees, reminds me of [NAME] before they bit the dust & Neutral & Concern, Reminiscence, Foreboding \\
Refrigerators? That's cool & Neutral & Amusement, Pun-Intended \\
I'm not sure what you mean by 18th century family here but otherwise thanks. & Confusion, Gratitude & Neutral \\
Thanks for the link. She actually likes some of the stuff! & Gratitude & Appreciation, Happiness \\
It sounds like you’re the one who is afraid of the internet. Relax, bud. You’re on r/cringe & Fear & Condescension, Annoyance \\
Forget new buildings, I work at the Chase Plaza building and am a little bummed that Seagram Tower made it but this didn’t. & Disappointment, Neutral & Bemused, A Little Bummed \\
\bottomrule
\end{tabular}
\caption{Ten random samples selected from GoEmotions training set, comparing human-annotated emotion labels and GPT-4 generated descriptive labels.}
\label{tab:GPT-4-vs-human}
\end{table}

\newpage

\section{Full Emotion Neighbor Retrieval Results}
\label{app:neighbor}

\begin{table}[H]
\centering
\small
\resizebox{1.05\textwidth}{!}{%
\begin{tabular}{l|p{4.6cm}|p{4.6cm}|p{4.6cm}}
\toprule
\textbf{Target Emotion} & \textbf{BERT (768D)} & \textbf{Ours (768D)} & \textbf{Ours (50D)} \\
\midrule
Admiration & Sympathy, Gratitude, Satisfaction, Reluctant Admiration & Reverence, Amazement, Awe, Attraction & Accomplishment, Reverence, Amazement, Pride \\
Gratitude & Satisfaction, Admiration, Affection, Jealousy & Appreciation, Grateful, Thankful, Politeness & Appreciation, Grateful, Thankful, Recognition \\
Approval & Disapproval, Recommendation, Fairness, Acceptance & Positive Surprise, Positive, Validation, Recommendation & Positive Interest, Positive Surprise, Positive, Favorability \\
Annoyance & Distraction, Reluctance, Frustrated, Avoidance & Irritation, Annoyed, Irritated, Exasperation & Irritation, Annoyed, Reluctance, Irritated \\
Curiosity & Suspicion, Surprise, Aggression, Distraction & Interest, Intrigue, Speculation, Curious & Interest, Slight Hopefulness, Curious, Fascination \\
Disapproval & Annoyance, Reluctance, Welcoming, Dismay & Judgment, Criticism, Condemnation, Dismay & Negative Opinion, Judgment, Criticism, Disagreement \\
Amusement & Laughter, Reluctant Admiration, Mock Frustration, Amused & Mild Amusement, Lack of Amusement, Lightheartedness, Mild Sarcasm & Mild Amusement, Lack of Amusement, Mock Frustration, Mock Seriousness \\
Love & Passion, Joy, Happiness, Pain & Affection, Passion, Adoration, Fondness & Affection, Joy, Good Wishes, Lust \\
Anger & Rage, Jealousy, Resentment, Contempt & Rage, Hostility, Hatred, Outrage & Rage, Hostility, Indignation, Hatred \\
Optimism & Cautious Optimism, Calmness, Hopefulness, Reassurance & Hopefulness, Hope, Optimistic, Hopeful & Hopefulness, Hope, Cautious Optimism, Hopeful \\
Joy & Hope, Grief, Calm, Celebration & Happiness, Celebration, Delight, Joyful & Happiness, Adoration, Celebration, Excitement \\
Sadness & Grief, Sad, Heartbreak, Disappointment & Sorrow, Heartbreak, Sad, Grief & Melancholy, Sad, Sorrow, Heartbreak \\
Confusion & Horror, Resentment, Jealousy, Impatience & Seeking Clarification, Seeking Help, Seeking Assistance, Seeking Advice & Confused, Awkwardness, Historical Trauma, Puzzlement \\
Disappointment & Regret, Frustration, Guilt, Relief & Dissatisfaction, Regret, Mild Disappointment, Disappointed & Dissatisfaction, Regret, Disappointed, Heartbreak \\
Realization & Disbelief, Distraction, Shock, Desperation & Surprise, Reflection, Denial, Acknowledgment & Denial, Shock, Focus, Shame \\
Surprise & Curiosity, Astonishment, Surprised, Realization & Realization, Astonishment, Mild Surprise, Shock & Mild Surprise, Shock, Astonishment, Realization \\
Caring & Denial, Kindness, Compassion, Protective & Empathetic, Considerate, Reassuring, Supportive & Considerate, Self-Assured, Affectionate, Accepting \\
Disgust & Irritation, Disgusted, Shame, Paranoia & Loathing, Disgusted, Horror, Hatred & Outrage, Disgusted, Condemnation, Dismay \\
Excitement & Delight, Anticipation, Panic, Astonishment & Joy, Delight, Enthusiasm, Triumph & Joy, Delight, Enthusiasm, Laughter \\
Desire & Lust, Jealousy, Longing, Affection & Lust, Longing, Craving, Eagerness & Longing, Lust, Passion, Misery \\
Fear & Panic, Distrust, Dread, Worry & Anxiety, Dread, Terror, Panic & Anxiety, Dread, Terror, Alarm \\
Remorse & Regret, Apology, Compassion, Guilt & Regret, Apology, Regretful, Apologetic & Regret, Apology, Regretful, Guilt \\
Embarrassment & Humiliation, Guilt, Concern, Frustration & Guilt, Humiliation, Self-Irritation, Shame & Guilt, Humiliation, Discomfort, Aversion \\
Nervousness & Tiredness, Helplessness, Awkwardness, Eagerness & Anxiety, Fear of Embarrassment, Numbness, Tiredness & Numbness, Desire for Emotional Relief, Fear of Embarrassment, Anxiety \\
Pride & Jealousy, Satisfaction, Empathy, Affection & Accomplishment, Confidence, Loyalty, Triumph & Reverence, Accomplishment, Belief, Authenticity \\
Relief & Relieved, Disappointment, Irritation, Grief & Gladness, Relieved, Glad, Satisfaction & Good, Relieved, Glad, Gladness \\
Grief & Heartbreak, Sadness, Desperation, Relief & Sorrow, Heartbreak, Sadness, Misery & Sorrow, Heartbreak, Sadness, Pain \\
\bottomrule
\end{tabular}
}
\caption{Top-4 most similar emotion terms retrieved for each target emotion using cosine similarity, using the BERT encoder, our contrastively distilled emotion model (768D), and its reduced 50D version.}
\label{tab:appendix-emotion-retrieval}
\end{table}

\end{document}